\documentclass[10pt,twocolumn,letterpaper]{article}

\usepackage[]{cvpr}

\usepackage[dvipsnames]{xcolor}

\definecolor{cvprblue}{rgb}{0.21,0.49,0.74}
\usepackage[pagebackref,breaklinks,colorlinks,citecolor=cvprblue]{hyperref}

\usepackage{indentfirst}
\usepackage{multicol}
\usepackage{multirow}
\usepackage{hyperref}
\usepackage{times}  
\usepackage{xcolor}
\hypersetup{
    linkcolor=blue        
}

\def\paperID{22565} 
\def\confName{CVPR}
\def\confYear{2026}

\title{Real-IAD MVN: A Multi-View Normal Vector Dataset and Benchmark for High-Fidelity Industrial Anomaly Detection}
\author{
    Wenbing Zhu$^{1,4\dag}$, Jianing Liang$^{3\dag}$, Linjie Cheng$^1$, Yurui Pan$^1$, \\
    Zhuhao Chen$^1$, Qingwang Yan$^1$, Yudong Cheng$^1$, Jianghui Zhang$^2$, Mingmin Chi$^1$, Bo Peng$^{2*}$
    \\
    \small $^1$Fudan University \hspace{2em} $^2$Shanghai Ocean University \hspace{2em} $^3$Donghua University \hspace{2em} $^4$Rongcheer Co., Ltd.
    \\
    {\tt \small \{wbzhu23, ljcheng24, yrpan24\}@m.fudan.edu.cn, 231310103@mail.dhu.edu.cn,}
    \\
    {\tt \small \{zhuhaochen24, qwyan25, 25113050248, mmchi\}@m.fudan.edu.cn, m240751960@st.shou.edu.cn,}
    \\
    {\tt \small bpeng@shou.edu.cn}
}
\vspace{-2em}
\begin{document}
\maketitle
\begingroup
\renewcommand\thefootnote{}
\footnotetext{$\dag$ Equal contribution. $*$ Corresponding author.}
\endgroup
\begin{abstract}
Industrial Anomaly Detection (IAD) is critical for quality control, but existing methods struggle with subtle, geometric defects. Standard 2D (RGB) images are sensitive to texture and lighting but often miss fine geometric anomalies. While 3D point clouds capture macro-shape, they are typically too sparse to detect micro-defects like scratches or pits.
We address this fundamental data limitation by introducing Real-IAD-MVN (Multi-View Normal), a large-scale industrial dataset. By upgrading our acquisition system, Real-IAD-MVN captures high-fidelity surface normal maps from five distinct viewpoints, replacing sparse 3D data entirely. This provides a comprehensive geometric representation at a micro-detail level, making previously invisible side-wall and occluded defects explicitly detectable.
Our experiments, conducted on this new dataset, first provide evidence that incorporating dense, multi-view pseudo-3D (surface normals) yields significantly better detection performance than using sparse 3D point cloud data. To further validate the dataset and provide a strong benchmark, we introduce a baseline method based on reconstruction, which learns to extract cross-modal unified prototypes from the image and normal map streams. We demonstrate that this unified prototype approach surpasses existing state-of-the-art multimodal fusion methods, highlighting the rich potential of our new dataset for advancing geometric anomaly detection.
\end{abstract}

\section{Introduction}

Unsupervised anomaly detection (AD) is a cornerstone of modern industrial manufacturing, ensuring product quality and safety by identifying defects without requiring vast datasets of failure examples \cite{bergmann2019mvtec,zou2022spot}. The primary goal is to model the distribution of normal products and then identify any deviation as an anomaly, both at the image level (classification) and pixel level (segmentation).

Initial AD methods relied on 2D (RGB) images \cite{bergmann2019mvtec, zou2022spot}. However, these approaches face a critical limitation: they are adept at finding *textural* or *color* anomalies but are fundamentally weak at identifying purely *geometric* defects, such as fine scratches, dents, or deformations. An anomaly's 2D appearance can be masked by challenging illumination, low-contrast, or reflective surfaces.

This weakness spurred the development of multimodal datasets, most notably those combining RGB with 3D data. Datasets like MVTec 3D-AD \cite{mvtec3d} and Real3D-AD \cite{liu2024real3d} utilize 3D point clouds to capture the macro-structural shape of objects. While this helps detect large-scale geometric anomalies (e.g., missing parts, severe warping), 3D point cloud data is often sparse, noisy, and at a resolution far too coarse to register the micro-geometric defects that are critical in high-precision manufacturing.

\begin{figure*}[t]
\begin{center}
\includegraphics[width=0.85\linewidth]{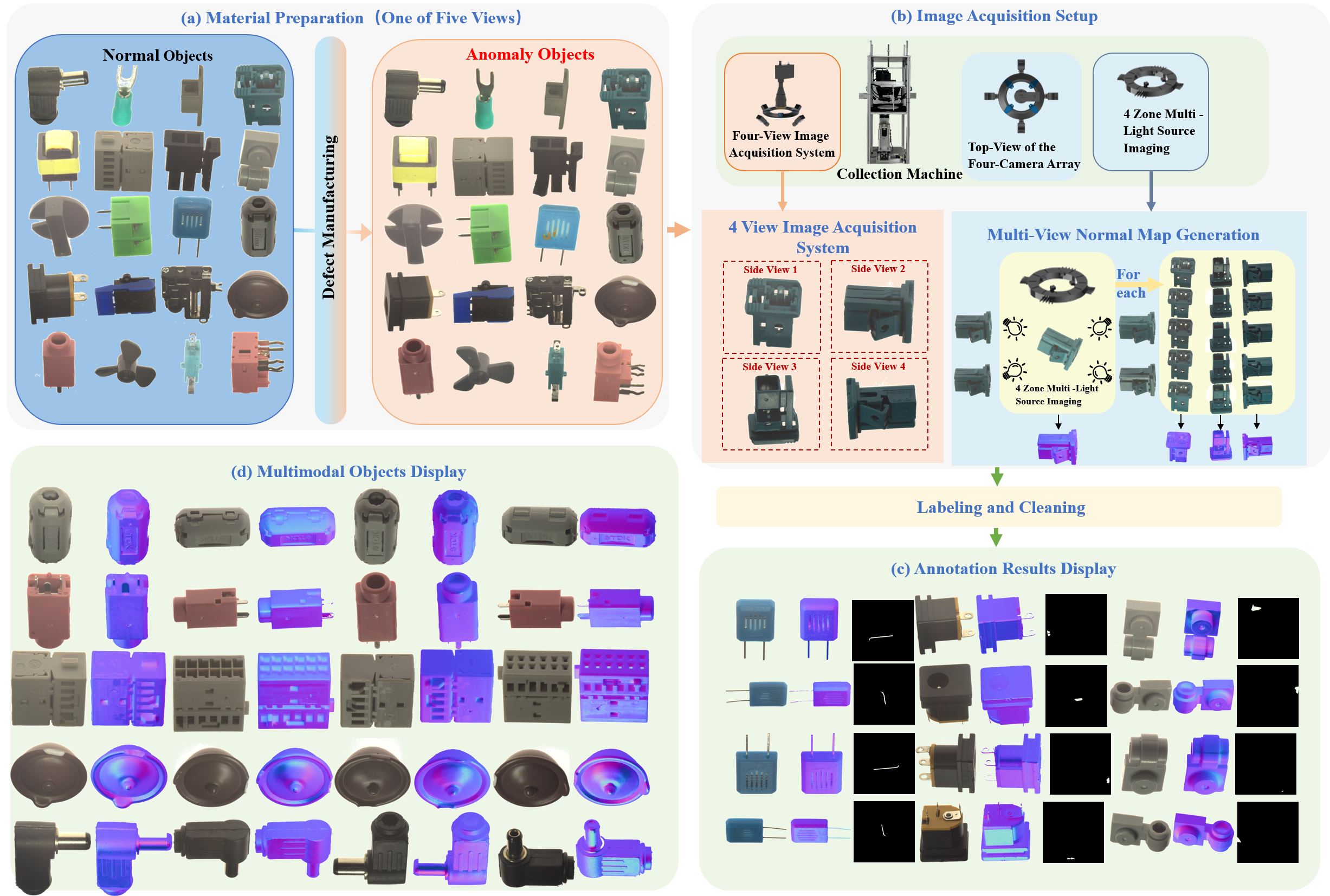}
\end{center}
\caption{Overview of the Real-IAD-MVN dataset acquisition and annotation pipeline. (a) Material preparation, showing examples of normal and anomalous objects from one of the five views. (b) The integrated multi-view photometric stereo (MVPS) acquisition gantry. (c) Examples of the pixel-perfect annotation masks and cleaning process. (d) A display of the final multimodal data, showing various RGB objects and their corresponding high-fidelity pseudo-3D (normal map) representations.}\label{fig:teaser}
\end{figure*}

A more promising direction for capturing fine geometry is photometric stereo (PS), which reconstructs high-fidelity surface normal maps \cite{li2025light}. Surface normals provide a dense, per-pixel representation of local geometry, making them extremely sensitive to subtle surface changes. Our prior work, Real-IAD D3 \cite{zhu2025real}, pioneered this by introducing a top-down pseudo-3D (normal map) modality. This addition proved highly effective for detecting defects on surfaces facing the camera. However, its single, top-down viewpoint remained a critical bottleneck. As illustrated in Figure~\ref{fig:teaser}, a defect on the side of an object is geometrically invisible to a top-down sensor, rendering it undetectable.

In this paper, we directly address this fundamental limitation. We present \textbf{Real-IAD-MVN}, a significant evolution of our previous work, detailed in Section~\ref{sec:dataset}. We have engineered a new acquisition system that captures high-fidelity photometric stereo data from \textbf{five calibrated viewpoints} (e.g., top, front, back, left, right). This multi-view normal (MVN) data provides a complete, 360-degree geometric profile of each sample, effectively eliminating the blind spots of single-view systems. We have extended the 20 categories from Real-IAD D3 \cite{zhu2025real}, creating a large-scale, challenging benchmark where view-dependent geometric anomalies are now the focus.

A new dataset demands validation via a strong baseline. We propose a reconstruction-based model designed to leverage this rich, multi-modal, multi-view data. Inspired by the challenge of cross-modal inconsistencies \cite{wang2024m3dm}, our baseline learns to extract \textbf{cross-modal unified prototypes} from the fused image and normal map streams. These prototypes holistically represent the joint "appearance-geometry" distribution of a normal sample. By tasking the model with reconstructing both modalities from this shared prototype representation, anomalies that violate this learned consistency (e.g., correct appearance but incorrect geometry) fail reconstruction and are thus clearly identified.

We validate our contributions comprehensively. First, experiments on our new dataset provide clear evidence that incorporating dense, multi-view pseudo-3D (surface normals) yields significantly better detection performance than using sparse 3D point cloud data. Second, we demonstrate that our unified prototype baseline surpasses existing state-of-the-art multimodal fusion methods, highlighting the rich potential of our new dataset.


\section{Related Work}

\subsection{Unsupervised Anomaly Detection}
Unsupervised AD methods are typically trained only on normal samples. These methods can be broadly categorized into three families.
\textbf{Embedding-based} methods learn a descriptive embedding for normal features. PaDiM \cite{defard2021padim} models patch features with a Gaussian distribution, while PatchCore \cite{patchcore} uses a coreset-based memory bank of normal features, achieving state-of-the-art performance with high efficiency.
\textbf{Reconstruction-based} methods train models like Autoencoders \cite{bergmann2018improving} or GANs \cite{schlegl2017unsupervised} to reconstruct normal samples, assuming anomalies will have high reconstruction error. Recent works like UniAD \cite{you2022unified} and OneNIP \cite{gao2024onenip} have successfully used Transformers for unified, multi-class AD, with OneNIP pioneering the use of a normal image as a prompt.
\textbf{Discriminator-based} methods, or "pseudo-anomaly" methods, generate synthetic anomalies and train a discriminative model to distinguish them from real normal samples. DRAEM \cite{zavrtanik2021draem} uses Perlin noise, CutPaste \cite{li2021cutpaste} pastes image patches, and SimpleNet \cite{liu2023simplenet} adds noise in the feature space.

\subsection{Multimodal Anomaly Detection}
Recognizing the limitations of 2D data, researchers have moved to multimodal AD.
\textbf{RGB + 3D Point Cloud:} The MVTec 3D-AD dataset \cite{mvtec3d} and Real3D-AD \cite{liu2024real3d} provided benchmarks for fusing 2D images with 3D point cloud data. Methods like M3DM \cite{wang2024m3dm,wang2025m3dm} proposed hybrid fusion techniques to leverage both surface texture (from 2D) and macro-shape (from 3D). However, as noted, point clouds are often too sparse for micro-defects.
\textbf{RGB + Pseudo-3D (Normals):} Our previous work, Real-IAD D3 \cite{zhu2025real}, was the first to incorporate high-precision, top-down photometric stereo (PS) data. This demonstrated the power of surface normals for detecting fine geometric defects. LINO UniPS \cite{li2025light} further confirmed the utility of PS for surface reconstruction, although not specifically for multi-view AD. Our new work builds directly on this, addressing the key limitation of single-viewpoint PS.

\begin{figure*}[h]
  \centering
  \includegraphics[width=0.8\textwidth]{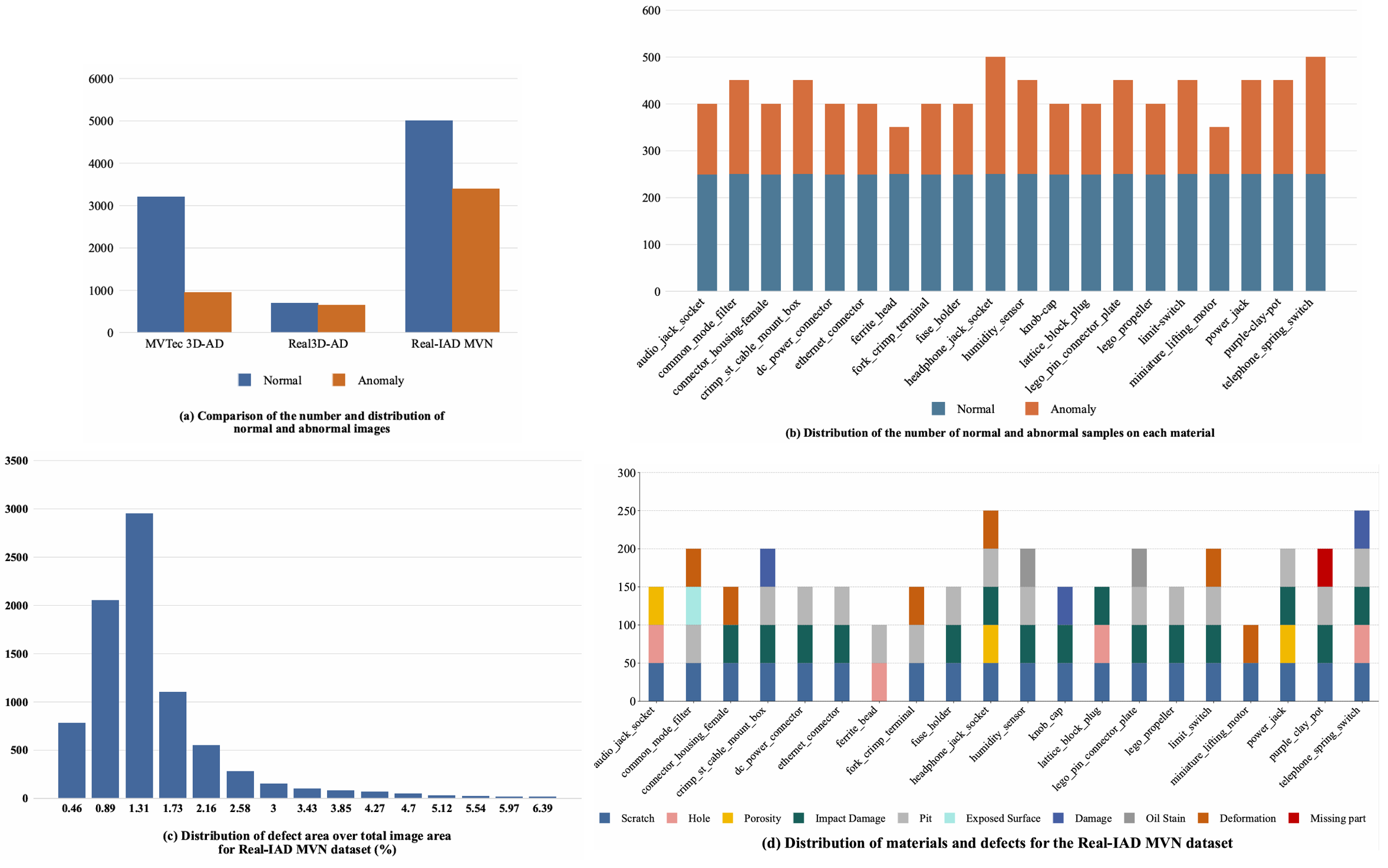}
  \caption{Statistical overview of the Real-IAD MVN dataset in comparison to MVTec 3D-AD and Real3D-AD, illustrating sample counts, normal/anomalous distribution per product, defect area ratios, and defect distribution patterns.}
\label{fig3}
\end{figure*}
\begin{table*}[h]
  \centering
  \caption{Comparison of Real-IAD-MVN with existing multimodal industrial anomaly detection datasets.}
  \label{tab:dataset_comparison}
  \resizebox{1.0\linewidth}{!}{
  \begin{tabular}{l|ccccc|cc}
    \toprule
    Dataset & Categories & Defect Types & Samples & Modalities & Geometric Resolution & Multi-View & Primary Geometric Data \\
    \midrule
    MVTec 3D-AD \cite{mvtec3d} & 10 & 33 & 4147 & RGB + 3D Point Cloud & $\sim$0.11mm (point precision) & No & Sparse Point Cloud \\
    Real3D-AD \cite{liu2024real3d} & 12 & 40 & 1254 & RGB + 3D Point Cloud & $\sim$0.01mm (point precision) & No & Sparse Point Cloud \\
    Real-IAD D3 \cite{zhu2025real} & 20 & 69 & 8450 & RGB + 3D Point Cloud + Pseudo-3D (Normal) & $\sim$0.002mm (normal precision) & No (Top-view normal) & Sparse PC + Single-view Normal \\
    \textbf{Real-IAD-MVN (Ours)} & \textbf{20} & \textbf{69} & \textbf{8450} & \textbf{MV-RGB + MV-Normal} & \textbf{$\sim$0.002mm (normal precision)} & \textbf{Yes (5 views)} & \textbf{Dense Multi-View Normals} \\
    \bottomrule
  \end{tabular}
  }
\end{table*}

\section{The Real-IAD-MVN Dataset}
\label{sec:dataset}

Our new dataset, Real-IAD-MVN, represents a paradigm shift in data acquisition for IAD, moving from single-view or sparse 3D data to dense, comprehensive multi-view surface geometry. It is built upon the established foundation of Real-IAD D3 \cite{zhu2025real} but introduces a fundamentally richer geometric modality.

\subsection{Multi-View Acquisition System}
We designed and constructed an integrated multi-view photometric stereo (MVPS) gantry. This system is crucial for capturing the novel data modality.
\begin{itemize}
    \item \textbf{Cameras:} A single high-resolution industrial camera ($3648 \times 5472$ pixels, similar to \cite{zhu2025real}) is mounted on a robotic arm capable of precisely positioning it at five pre-calibrated viewpoints around the object (e.g., top, front, back, left, right).
    \item \textbf{Photometric Stations:} Co-located with each camera viewpoint is a photometric stereo station. Each station employs multiple (e.g., 4 or more) directional light sources positioned at known angles relative to the camera \cite{zhu2025real, li2025light}.
    \item \textbf{Synchronized Capture:} For each sample, the system captures a sequence of images under different lighting directions from each of the five viewpoints. Additionally, a standard diffusely-lit RGB image ($I_v$) is captured from each view.
\end{itemize}
This setup ensures precise alignment between the RGB image and the computed normal map for each view, and the calibration between views allows for potential 3D reconstruction or cross-view warping if needed, though our proposed method operates primarily in 2D feature space.

\subsection{Photometric Stereo for Normals}
From the sequence of images captured under varying illumination at each viewpoint $v$, we compute the surface normal map $N_v$. Assuming Lambertian reflectance for simplicity (though more complex models can be used \cite{li2025light}), the intensity $I(x, y)$ at a pixel $(x, y)$ under a known light source direction $L \in \mathbb{R}^3$ is related to the surface normal $n(x, y) \in \mathbb{R}^3$ and albedo $\rho(x, y)$ by:
$$ I(x, y) = \rho(x, y) \max(0, L \cdot n(x, y)) $$
With at least three non-collinear light source directions $\{L_k\}_{k=1}^K$ ($K \ge 3$) and their corresponding intensities $\{I_k(x, y)\}_{k=1}^K$, we can solve for the scaled normal vector $g(x, y) = \rho(x, y) n(x, y)$ using a least-squares approach \cite{woodham1980photometric, zhu2025real}. Let $\mathbf{I}$ be the $K \times 1$ vector of intensities and $\mathbf{L}$ be the $K \times 3$ matrix of light directions. Then:
$$ \mathbf{I} = \mathbf{L} g $$
$$ g = (\mathbf{L}^T \mathbf{L})^{-1} \mathbf{L}^T \mathbf{I} $$
The final normal $n(x, y)$ is obtained by normalizing $g(x, y)$: $n = g / ||g||_2$. This process is repeated for each viewpoint $v$, yielding the set $\{N_v\}_{v=1}^5$. The resulting normal maps offer micrometer-level precision in capturing surface geometry, far exceeding the detail available from typical 3D scanners used in prior datasets \cite{zhu2025real}. 

\subsection{Dataset Statistics and Comparison}
Real-IAD-MVN retains the 20 object categories and 69 defect types from Real-IAD D3 \cite{zhu2025real}. The total number of samples remains 8,450 (5,000 normal, 3,450 abnormal). Each sample now consists of:
\begin{itemize}
    \item 5 RGB Images $\{I_v\}_{v=1}^5$
    \item 5 Surface Normal Maps $\{N_v\}_{v=1}^5$
    \item Pixel-level anomaly masks $\{M_v\}_{v=1}^5$ (where available for abnormal samples)
\end{itemize}
This constitutes a dataset size roughly $5 \times$ larger than Real-IAD D3 in terms of image/map count. The critical difference is the replacement of the single top-down normal map and sparse 3D point cloud with five high-fidelity normal maps, enabling comprehensive geometric analysis.

Table~\ref{tab:dataset_comparison} compares Real-IAD-MVN with prior multimodal IAD datasets. Our key innovations are the \textbf{multi-view capture} and the reliance on \textbf{dense surface normals} as the primary geometric modality, offering superior resolution and coverage compared to point clouds or single-view systems.

\section{Methodology: A Unified Prototype Baseline}
\label{sec:method}

To validate the descriptive power of our new Real-IAD-MVN dataset, we establish a strong baseline model, the \textbf{Cross-modal Prototype Reconstruction Network (CPRN)}. This model is designed to leverage the multi-modal (RGB and Normal Vector) data from a single viewpoint, as our dataset's primary comparison is against single-view 2D+3D methods which require strict alignment. The core hypothesis, inspired by recent work on intrinsic prototype learning~\cite{lee2023inpformer, INP-Former}, is that a robust model of normality must capture the \textit{joint probability distribution} $P(\text{appearance}, \text{geometry})$.

\begin{figure}[t]
\begin{center}
\includegraphics[width=\linewidth]{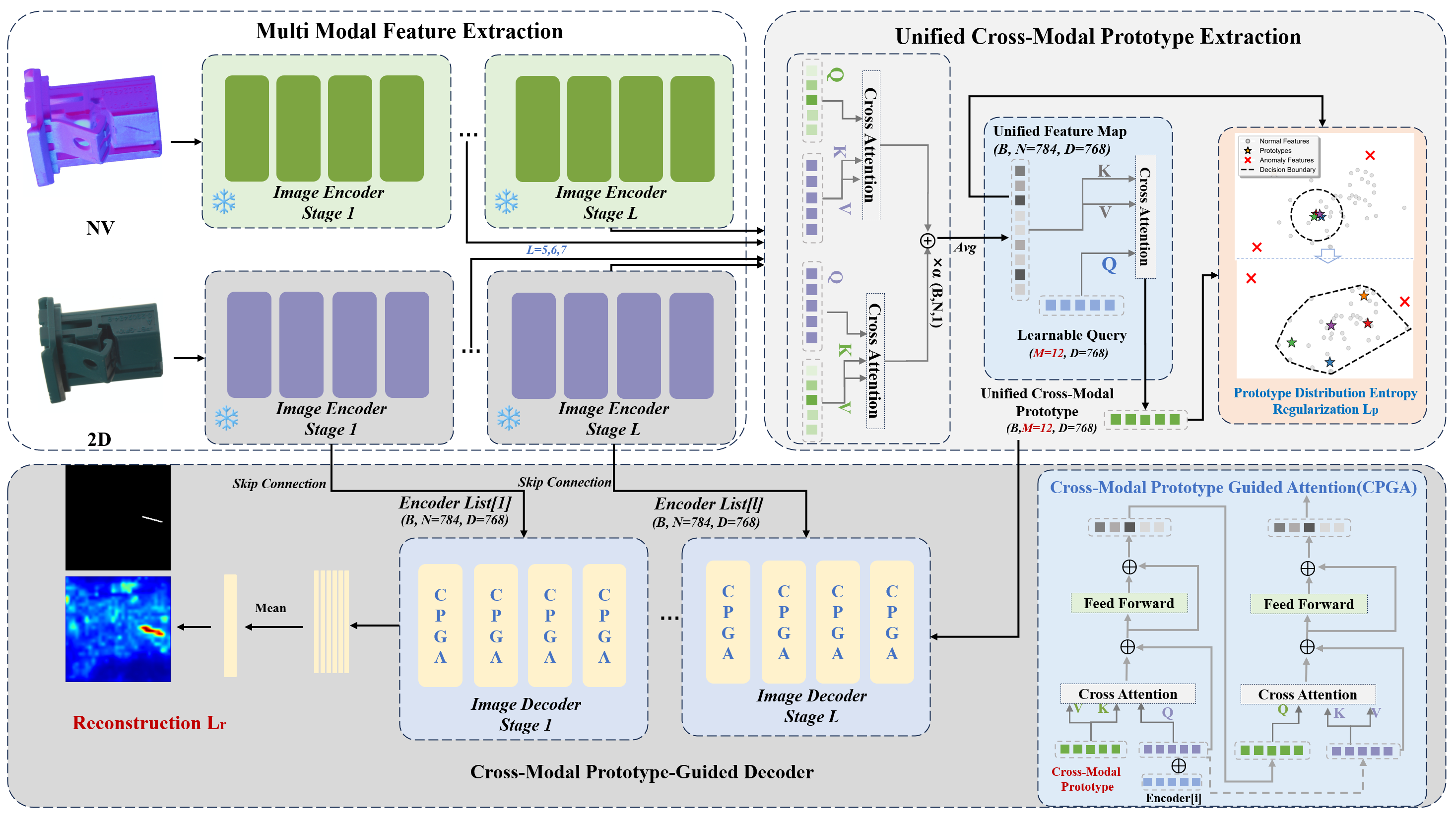} 
\end{center}
   \caption{Overview of our CPRN baseline architecture. Multi-modal features (NV and 2D) are extracted by frozen encoders. The UCP module learns a set of unified prototypes $P_{ucmp}$ by attending to both feature streams. The decoder then reconstructs the 2D features using $P_{ucmp}$ as a shared dictionary, guided by skip connections and CPGA blocks. Anomalies are detected via high reconstruction loss ($L_r$), and the prototype space is regularized by an entropy loss ($L_p$).}
\label{fig:architecture}
\end{figure}

Our baseline is a reconstruction-based network, illustrated in Figure \ref{fig:architecture}. It learns a set of Unified Cross-Modal Prototypes (UCMPs) that represent holistic, normal "concepts" (e.g., "this specific shiny, flat surface"). Anomalies are detected as features that cannot be well-reconstructed from this learned dictionary of normality. The network consists of three main stages: (1) Multi-Modal Feature Extraction, (2) Unified Cross-Modal Prototype Extraction (UCP), and (3) Cross-Modal Prototype-Guided Reconstruction.

\subsection{Multi-Modal Feature Extraction}
Given an input RGB image $I_{rgb} \in \mathbb{R}^{H \times W \times 3}$ and its corresponding surface normal (NV) map $I_{nv} \in \mathbb{R}^{H \times W \times 3}$, we first extract multi-level features. We use two parallel, pre-trained Vision Transformer (ViT) encoders, $E_{rgb}$ and $E_{nv}$, with frozen weights~\cite{patchcore, defard2021padim}. The encoders produce multi-level feature sequences:
$$E_{rgb} = \{E_{rgb,l}\}_{l=1}^{L} \quad \text{and} \quad E_{nv} = \{E_{nv,l}\}_{l=1}^{L}$$
where $l$ is the layer index, $N$ is the number of patch tokens, and $D$ is the feature dimension. We denote the final-layer features as $F_{rgb} = E_{rgb,L}$ and $F_{nv} = E_{nv,L}$.

\subsection{Unified Cross-Modal Prototype Extraction (UCP)}
The UCP module creates a single, compact set of $M$ prototypes representing the joint distribution. We initialize $M$ learnable tokens $Q_{learn} \in \mathbb{R}^{M \times D}$, which act as queries. As shown in Figure \ref{fig:architecture} (top-right), these queries attend to both feature streams:
\begin{equation}
    P'_{rgb} = \text{Attention}(Q_{learn}, F_{rgb}, F_{rgb})
\end{equation}
\begin{equation}
    P'_{nv} = \text{Attention}(Q_{learn}, F_{nv}, F_{nv})
\end{equation}
The aggregated information is averaged and refined via an FFN to produce the final Unified Cross-Modal Prototypes $P_{ucmp} \in \mathbb{R}^{M \times D}$:
\begin{equation}
    P_{ucmp} = \text{FFN}\left(\text{LayerNorm}\left(Q_{learn} + \frac{1}{2}(P'_{rgb} + P'_{nv})\right)\right)
\end{equation}
This set $P_{ucmp}$ serves as a unified dictionary of all valid appearance-geometry concepts in the normal sample.

\subsection{Cross-Modal Prototype-Guided Reconstruction}
The final stage uses the learned prototypes $P_{ucmp}$ to reconstruct the original RGB features, forcing the reconstruction to be consistent with the joint prototype concepts. The decoder consists of $L$ stages, mirroring the encoder.

Each decoder stage $l$ employs a Cross-Modal Prototype Guided Attention (CPGA) block. As shown in Figure \ref{fig:architecture} (bottom-right), this block takes the corresponding skip-connection feature $E_{rgb,l}$ as the Query (Q). The Key (K) and Value (V) are the \textit{same} set of Unified Cross-Modal Prototypes $P_{ucmp}$.
\begin{equation}
    F'_{dec,l} = \text{Attention}(E_{rgb,l}, P_{ucmp}, P_{ucmp})
\end{equation}
The output $F'_{dec,l}$ is the "reconstructed" version of $E_{rgb,l}$ using only the normal concepts available in $P_{ucmp}$. This output is then refined with a residual connection and an FFN. If an input contains an anomaly (e.g., normal color but anomalous geometry), the $P_{ucmp}$ set will be "confused" and will not contain the correct concepts to reconstruct the RGB features, leading to a high reconstruction error.

\subsection{Optimization Objective}
The network is trained end-to-end with a composite loss function:
$L_{total} = L_{r} + \lambda_p L_{p}$,
where $\lambda_p$ is a balancing hyperparameter.

\noindent\textbf{Reconstruction Loss ($L_{r}$)}. This primary loss measures the fidelity of the reconstruction. As shown in Figure \ref{fig:architecture}, we compute the mean of all skip connections from the RGB encoder ($F_{rgb, \text{mean}}$) and the mean of all outputs from the decoder ($\tilde{F}_{rgb, \text{mean}}$). The loss is the mean cosine distance between them:
\begin{equation}
    L_{r} = 1 - \text{sim}(F_{rgb, \text{mean}}, \tilde{F}_{rgb, \text{mean}})
\end{equation}
where $\text{sim}(\cdot)$ denotes cosine similarity.

\noindent\textbf{Prototype Distribution Entropy Regularization ($L_{p}$)}. This regularization loss, also depicted in the UCP module diagram, prevents "prototype collapse" by encouraging the model to utilize all $M$ prototypes equally. It computes the negative entropy of the prototype assignment distribution $q$, thereby maximizing diversity:
\begin{equation}
    L_{p} = \sum_{j=1}^{M} q_j \log(q_j + \epsilon)
\end{equation}
This ensures the learned prototype space is well-structured.

\noindent\textbf{Inference.} At test time, an image-level anomaly score is computed from $L_{r}$. A pixel-level anomaly map is generated by computing the patch-wise cosine distance between the final layer encoder features $F_{rgb}$ and their corresponding reconstructions from the final decoder layer $\tilde{F}_{rgb,L}$.
\begin{table*}[htbp]
  \centering
  \caption{Performance comparison of different multimodal anomaly detection methods on the \textbf{Real-IAD-MVN (Ours)} dataset. \textbf{Crucially, this table evaluates a single-view (top-down) scenario} to ensure a fair comparison with traditional 2D+3D methods that require strict alignment. The table is grouped by modality: Single Modality (RGB), traditional multimodal (2D+3D), and our proposed modality (RGB+NV). Our baseline, CPRN achieves new state-of-the-art results against all single-view competitors.}
    \resizebox{\textwidth}{!}{%
    \begin{tabular}{l|ccc|ccc|ccc|ccc|ccc|ccc}
        \toprule
    \textbf{Modality} & \multicolumn{6}{c|}{\textbf{RGB}} & \multicolumn{9}{c|}{\textbf{2D+3D}} & \multicolumn{3}{c}{\textbf{NV+RGB}} \\
    \cmidrule(r){2-7} \cmidrule(r){8-16} \cmidrule(r){17-19}
    \textbf{Model} & \multicolumn{3}{c|}{\textbf{SimpleNet}} & \multicolumn{3}{c|}{\textbf{Cflow}} & \multicolumn{3}{c|}{\textbf{AST}} & \multicolumn{3}{c|}{\textbf{PatchCore}} & \multicolumn{3}{c|}{\textbf{M3DM}} & \multicolumn{3}{c}{\textbf{CPRN (Ours)}} \\
    \midrule
    \textbf{Metrics} & \textbf{I-Auroc} & \textbf{P-AUROC} & \textbf{P-AUPRO} & \textbf{I-Auroc} & \textbf{P-AUROC} & \textbf{P-AUPRO} & \textbf{I-Auroc} & \textbf{P-AUROC} & \textbf{P-AUPRO} & \textbf{I-Auroc} & \textbf{P-AUROC} & \textbf{P-AUPRO} & \textbf{I-Auroc} & \textbf{P-AUROC} & \textbf{P-AUPRO} & \textbf{I-Auroc} & \textbf{P-AUROC} & \textbf{P-AUPRO} \\
    \midrule
    audio\_jack\_socket & 0.973 & 0.926 & 0.710 & 0.943 & 0.944 & 0.725 & 0.860 & 0.590 & 0.410 & 0.926 & 0.673 & 0.505 & 0.981 & 0.699 & 0.550 & \textbf{0.986} & \textbf{0.979} & \textbf{0.878} \\
    common\_mode\_filter & 0.717 & 0.822 & 0.600 & 0.271 & 0.847 & 0.615 & 0.899 & 0.802 & 0.590 & 0.523 & 0.922 & 0.790 & 0.580 & 0.934 & 0.820 & \textbf{0.945} & \textbf{0.942} & \textbf{0.741} \\
    connector\_housing\_female & 0.795 & 0.891 & 0.690 & 0.839 & 0.921 & 0.705 & 0.914 & 0.716 & 0.510 & 0.870 & 0.919 & 0.785 & 0.920 & 0.979 & 0.900 & \textbf{0.997} & \textbf{0.972} & \textbf{0.791} \\
    crimp\_st\_cable\_mount\_box & 0.372 & 0.745 & 0.510 & 0.180 & 0.442 & 0.300 & 0.485 & 0.589 & 0.390 & 0.713 & 0.931 & 0.810 & 0.749 & 0.933 & 0.830 & \textbf{0.995} & \textbf{0.959} & \textbf{0.771} \\
    dc\_power\_connector & 0.661 & 0.725 & 0.500 & 0.661 & 0.726 & 0.500 & 0.995 & 0.770 & 0.550 & 0.720 & 0.921 & 0.790 & 0.715 & 0.950 & 0.850 & \textbf{0.994} & \textbf{0.901} & \textbf{0.702} \\
    ethernet\_connector & 0.981 & 0.866 & 0.650 & 0.967 & 0.853 & 0.640 & 1.000 & 0.906 & 0.710 & 0.947 & 0.956 & 0.840 & 0.983 & 0.978 & 0.910 & \textbf{0.992} & \textbf{0.989} & \textbf{0.958} \\
    ferrite\_bead & 0.408 & 0.806 & 0.590 & 0.529 & 0.914 & 0.690 & 0.894 & 0.817 & 0.600 & 0.913 & 0.932 & 0.815 & 0.965 & 0.966 & 0.880 & \textbf{0.990} & \textbf{0.988} & \textbf{0.946} \\
    fork\_crimp\_terminal & 0.416 & 0.945 & 0.720 & 0.462 & 0.657 & 0.450 & 0.595 & 0.773 & 0.560 & 0.769 & 0.952 & 0.840 & 0.780 & 0.964 & 0.870 & \textbf{0.950} & \textbf{0.981} & \textbf{0.825} \\
    fuse\_holder & 0.564 & 0.957 & 0.750 & 0.853 & 0.861 & 0.640 & 0.597 & 0.754 & 0.540 & 0.736 & 0.927 & 0.800 & 0.770 & 0.948 & 0.850 & \textbf{0.996} & \textbf{0.994} & \textbf{0.980} \\
    headphone\_jack\_socket & 0.933 & 0.879 & 0.670 & 0.996 & 0.914 & 0.700 & 0.660 & 0.696 & 0.490 & 0.919 & 0.942 & 0.830 & 0.982 & 0.982 & 0.920 & \textbf{0.997} & \textbf{0.969} & \textbf{0.857} \\
    humidity\_sensor & 0.737 & 0.890 & 0.680 & 0.781 & 0.836 & 0.610 & 0.565 & 0.723 & 0.510 & 0.689 & 0.933 & 0.810 & 0.717 & 0.958 & 0.860 & \textbf{0.979} & \textbf{0.970} & \textbf{0.819} \\
    knob\_cap & 0.672 & 0.879 & 0.670 & 0.637 & 0.893 & 0.680 & 0.919 & 0.656 & 0.450 & 0.903 & 0.958 & 0.850 & 0.925 & 0.938 & 0.830 & \textbf{0.966} & \textbf{0.954} & \textbf{0.875} \\
    lattice\_block\_plug & 0.790 & 0.898 & 0.690 & 0.833 & 0.852 & 0.630 & 0.842 & 0.919 & 0.730 & 0.911 & 0.923 & 0.800 & 0.917 & 0.958 & 0.860 & \textbf{0.961} & \textbf{0.976} & \textbf{0.926} \\
    lego\_pin\_connector\_plate & 0.857 & 0.947 & 0.730 & 0.828 & 0.877 & 0.660 & 0.847 & 0.629 & 0.420 & 0.662 & 0.759 & 0.580 & 0.681 & 0.734 & 0.580 & \textbf{0.950} & \textbf{0.992} & \textbf{0.945} \\
    lego\_propeller & 0.939 & 0.799 & 0.580 & 0.615 & 0.739 & 0.510 & 0.471 & 0.703 & 0.500 & 0.540 & 0.727 & 0.560 & 0.530 & 0.773 & 0.610 & \textbf{0.998} & \textbf{0.952} & \textbf{0.806} \\
    limit\_switch & 0.823 & 0.790 & 0.570 & 0.846 & 0.950 & 0.740 & 0.804 & 0.641 & 0.430 & 0.822 & 0.938 & 0.820 & 0.863 & 0.966 & 0.890 & \textbf{0.979} & \textbf{0.969} & \textbf{0.864} \\
    miniature\_lifting\_motor & 0.402 & 0.760 & 0.530 & 0.402 & 0.799 & 0.570 & 0.766 & 0.467 & 0.310 & 0.948 & 0.962 & 0.870 & 0.975 & 0.991 & 0.940 & \textbf{0.923} & \textbf{0.967} & \textbf{0.703} \\
    power\_jack & 0.176 & 0.489 & 0.310 & 0.354 & 0.664 & 0.450 & 0.564 & 0.645 & 0.440 & 0.981 & 0.923 & 0.800 & 0.996 & 0.902 & 0.780 & \textbf{0.992} & \textbf{0.937} & \textbf{0.662} \\
    purple\_clay\_pot & 0.343 & 0.938 & 0.720 & 0.343 & 0.571 & 0.390 & 0.635 & 0.445 & 0.290 & 0.921 & 0.961 & 0.870 & 0.944 & 0.953 & 0.860 & \textbf{0.896} & \textbf{0.981} & \textbf{0.951} \\
    telephone\_spring\_switch & 0.627 & 0.916 & 0.700 & 0.575 & 0.910 & 0.690 & 0.951 & 0.551 & 0.380 & 0.827 & 0.944 & 0.830 & 0.856 & 0.936 & 0.830 & \textbf{0.975} & \textbf{0.976} & \textbf{0.905} \\
    \midrule
    \textbf{Avg} & 0.659 & 0.843 & 0.629 & 0.645 & 0.808 & 0.595 & 0.776 & 0.699 & 0.521 & 0.812 & 0.905 & 0.790 & 0.841 & 0.922 & 0.830 & \textbf{0.973} & \textbf{0.968} & \textbf{0.841} \\
    \bottomrule
    \end{tabular}%
    }
  \label{tab:main_results}%
\end{table*}

\begin{table*}[htbp]
  \centering
  \caption{Performance comparison of RGB+NV multimodal methods on the \textbf{full Real-IAD-MVN dataset (all 5 views)}. This experiment highlights the performance of different fusion strategies when leveraging the complete multi-view RGB and multi-view NV data. Our baseline, CPRN, sets a new state-of-the-art on this full dataset.}
    \resizebox{\textwidth}{!}{%
    \begin{tabular}{l|ccc|ccc|ccc|ccc|ccc|ccc}
        \toprule
    \textbf{Model} & \multicolumn{3}{c|}{\textbf{PatchCore}} & \multicolumn{3}{c|}{\textbf{M3DM}} & \multicolumn{3}{c|}{\textbf{D3M}} & \multicolumn{3}{c|}{\textbf{Dinomaly}} & \multicolumn{3}{c|}{\textbf{InpFormer}} & \multicolumn{3}{c}{\textbf{CPRN (Ours)}} \\
    \cmidrule(r){2-4} \cmidrule(r){5-7} \cmidrule(r){8-10} \cmidrule(r){11-13} \cmidrule(r){14-16} \cmidrule(r){17-19}
    \textbf{Category} & \textbf{I-AUROC} & \textbf{P-AUROC} & \textbf{P-AUPRO} & \textbf{I-AUROC} & \textbf{P-AUROC} & \textbf{P-AUPRO} & \textbf{I-AUROC} & \textbf{P-AUROC} & \textbf{P-AUPRO} & \textbf{I-AUROC} & \textbf{P-AUROC} & \textbf{P-AUPRO} & \textbf{I-AUROC} & \textbf{P-AUROC} & \textbf{P-AUPRO} & \textbf{I-AUROC} & \textbf{P-AUROC} & \textbf{P-AUPRO} \\
    \midrule
    dc\_power\_connector & 0.932 & 0.991 & 0.488 & 0.920 & 0.990 & 0.481 & 0.929 & 0.991 & 0.487 & 0.935 & 0.992 & 0.491 & 0.938 & 0.993 & 0.495 & \textbf{0.9965} & \textbf{0.9044} & \textbf{0.7117} \\
    fork\_crimp\_terminal & 0.843 & 0.963 & 0.769 & 0.835 & 0.961 & 0.758 & 0.832 & 0.962 & 0.765 & 0.838 & 0.965 & 0.771 & 0.841 & 0.966 & 0.775 & \textbf{0.9536} & \textbf{0.9845} & \textbf{0.8361} \\
    ethernet\_connector & 0.969 & 0.994 & 0.582 & 0.962 & 0.992 & 0.575 & 0.967 & 0.994 & 0.580 & 0.972 & 0.995 & 0.588 & 0.974 & 0.996 & 0.591 & \textbf{0.9954} & \textbf{0.9922} & \textbf{0.9684} \\
    fuse\_holder & 0.846 & 0.981 & 0.574 & 0.900 & 0.980 & 0.565 & 0.919 & 0.982 & 0.551 & 0.925 & 0.983 & 0.559 & 0.928 & 0.984 & 0.563 & \textbf{0.9983} & \textbf{0.9972} & \textbf{0.9901} \\
    common\_mode\_filter & 0.915 & 0.988 & 0.728 & 0.902 & 0.986 & 0.715 & 0.949 & 0.993 & 0.743 & 0.940 & 0.990 & 0.735 & \textbf{0.9500} & \textbf{0.994} & 0.748 & 0.9479 & 0.9455 & \textbf{0.7511} \\
    crimp\_st\_cable\_mount\_box & 0.985 & 0.998 & 0.570 & 0.980 & 0.997 & 0.562 & 0.985 & 0.998 & 0.594 & 0.988 & 0.999 & 0.601 & 0.990 & 0.999 & 0.605 & \textbf{0.9980} & \textbf{0.9629} & \textbf{0.7804} \\
    connector\_housing\_female & 0.995 & 0.999 & 0.610 & 0.993 & 0.998 & 0.595 & 0.999 & 0.999 & 0.602 & 0.999 & 0.999 & 0.605 & \textbf{1.000} & \textbf{1.000} & 0.608 & \textbf{1.0000} & 0.9752 & \textbf{0.8005} \\
    lego\_pin\_connector\_plate & 0.689 & 0.948 & 0.809 & 0.680 & 0.945 & 0.795 & 0.717 & 0.952 & 0.812 & 0.725 & 0.955 & 0.818 & 0.730 & 0.956 & 0.821 & \textbf{0.9526} & \textbf{0.9947} & \textbf{0.9564} \\
    knob\_cap & 0.907 & 0.977 & 0.609 & 0.898 & 0.975 & 0.601 & 0.921 & 0.981 & 0.617 & 0.925 & 0.982 & 0.622 & 0.928 & 0.983 & 0.625 & \textbf{0.9686} & \textbf{0.9572} & \textbf{0.8861} \\
    lattice\_block\_plug & 0.695 & 0.956 & 0.578 & 0.681 & 0.952 & 0.565 & 0.759 & 0.964 & 0.587 & 0.768 & 0.966 & 0.593 & 0.772 & 0.968 & 0.598 & \textbf{0.9640} & \textbf{0.9797} & \textbf{0.9364} \\
    humidity\_sensor & 0.844 & 0.973 & 0.617 & 0.835 & 0.971 & 0.609 & 0.855 & 0.975 & 0.619 & 0.860 & 0.976 & 0.624 & 0.863 & 0.977 & 0.627 & \textbf{0.9819} & \textbf{0.9732} & \textbf{0.8298} \\
    ferrite\_bead & 0.907 & 0.974 & 0.580 & 0.899 & 0.972 & 0.571 & 0.930 & 0.981 & 0.577 & 0.935 & 0.982 & 0.583 & 0.938 & 0.983 & 0.587 & \textbf{0.9922} & \textbf{0.9911} & \textbf{0.9560} \\
    power\_jack & 0.947 & 0.993 & 0.526 & 0.941 & 0.991 & 0.518 & 0.960 & 0.994 & 0.545 & 0.963 & 0.995 & 0.551 & 0.965 & 0.995 & 0.554 & \textbf{0.9951} & \textbf{0.9406} & \textbf{0.6729} \\
    telephone\_spring\_switch & 0.781 & 0.969 & 0.672 & 0.772 & 0.965 & 0.665 & 0.834 & 0.977 & 0.692 & 0.840 & 0.979 & 0.698 & 0.843 & 0.980 & 0.702 & \textbf{0.9783} & \textbf{0.9794} & \textbf{0.9162} \\
    miniature\_lifting\_motor & 0.856 & 0.937 & 0.604 & 0.847 & 0.933 & 0.595 & 0.850 & 0.927 & 0.638 & \textbf{0.9300} & 0.931 & 0.645 & 0.928 & \textbf{0.972} & 0.710 & 0.9262 & 0.9706 & \textbf{0.7141} \\
    purple\_clay\_pot & 0.833 & 0.966 & 0.686 & 0.825 & 0.964 & 0.678 & 0.838 & 0.969 & 0.707 & 0.843 & 0.971 & 0.713 & 0.845 & 0.972 & 0.716 & \textbf{0.8993} & \textbf{0.9843} & \textbf{0.9617} \\
    headphone\_jack\_socket & 0.970 & 0.997 & 0.555 & 0.963 & 0.995 & 0.546 & 0.987 & 0.997 & 0.531 & 0.990 & 0.998 & 0.535 & 0.992 & 0.998 & 0.538 & \textbf{0.9991} & \textbf{0.9722} & \textbf{0.8671} \\
    lego\_propeller & 0.993 & 0.999 & 0.562 & 0.990 & 0.998 & 0.553 & \textbf{1.000} & \textbf{1.000} & 0.572 & 0.998 & \textbf{1.000} & 0.575 & 0.999 & \textbf{1.000} & 0.578 & 0.9999 & 0.9557 & \textbf{0.8171} \\
    limit\_switch & 0.878 & 0.980 & 0.635 & 0.869 & 0.978 & 0.626 & 0.914 & 0.986 & 0.619 & 0.918 & 0.987 & 0.623 & 0.920 & 0.988 & 0.626 & \textbf{0.9824} & \textbf{0.9728} & \textbf{0.8748} \\
    audio\_jack\_socket & 0.922 & 0.992 & 0.570 & 0.913 & 0.990 & 0.562 & 0.931 & 0.993 & 0.571 & 0.935 & 0.994 & 0.576 & 0.938 & 0.995 & 0.580 & \textbf{0.9889} & \textbf{0.9815} & \textbf{0.8897} \\
    \midrule
    \textbf{Mean} & 0.885 & 0.978 & 0.616 & 0.878 & 0.975 & 0.608 & 0.904 & 0.981 & 0.620 & 0.913 & 0.983 & 0.630 & 0.916 & 0.984 & 0.633 & \textbf{0.9759} & \textbf{0.9707} & \textbf{0.8558} \\
    \bottomrule
    \end{tabular}%
    }
  \label{tab:main_results_multiview}%
\end{table*}

\section{Experiments}
\label{sec:experiments}

\subsection{Experimental Setup}

\noindent\textbf{Datasets and Metrics.}
Our primary experiments are conducted on our newly proposed \textbf{Real-IAD-MVN} dataset. We also perform a validation experiment on the public Real-IAD (MV-RGB) benchmark~\cite{wang2024real} to confirm our baseline's architectural robustness. We evaluate all methods using standard IAD metrics: Image-level AUROC (I-AUROC), Image-level F1-Score (I-F1), Pixel-level AUROC (P-AUROC), and Pixel-level AUPRO (P-AUPRO).

\noindent\textbf{Experimental Settings and Baselines.}
We validate our contributions with two key experiments on our \textbf{Real-IAD-MVN} dataset: (1) A \textbf{Single-View (Top-Down)} comparison to fairly test our PS-3D (RGB+Normal) modality against traditional 2D+3D (RGB+Point Cloud)~\cite{zhu2025real}. (2) A \textbf{Full Dataset (5-View)} comparison using our complete RGB + Normal data to test SOTA 2D+PS fusion methods. We compare against a comprehensive suite of baselines, including single-modality (SimpleNet~\cite{liu2023simplenet}, Cflow\cite{gudovskiy2022cflow}), 2D+3D (AST, PatchCore~\cite{patchcore}, M3DM~\cite{wang2024m3dm}), and 2D+PS methods (D3M~\cite{zhu2025real}, Dinomaly~\cite{dinomaly}, INP-Former~\cite{INP-Former}). We also validate our baseline on the Real-IAD (MV-RGB) benchmark~\cite{wang2024real}.

\begin{figure*}[t]
\begin{center}
\includegraphics[width=0.85\linewidth]{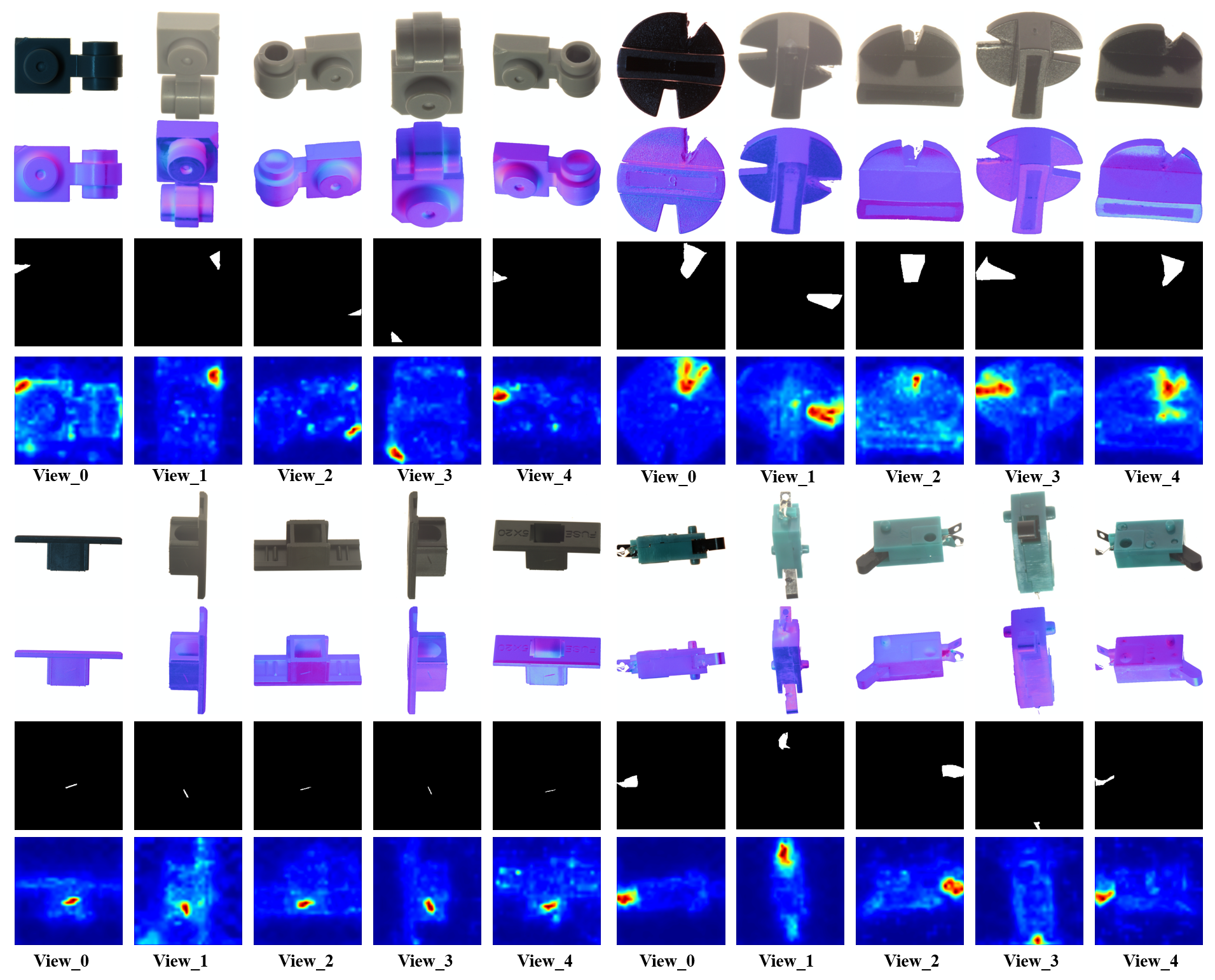} 
\end{center}
   \caption{Qualitative results of our single-view CPRN model applied independently to multiple views from the Real-IAD-MVN dataset. For two categories. For two categories (top: 'lego\_pin\_connector\_plate', bottom: 'limit\_switch'), we show the RGB image, corresponding PS normal map, ground truth mask, and our CPRN's predicted anomaly map from multiple viewpoints. Note how defects invisible in one view (e.g., top-down) become clearly visible from a side view, demonstrating the necessity of our multi-view normal dataset.}
\label{fig:multiview}
\end{figure*}

\begin{figure*}[h]
\begin{center}
\includegraphics[width=0.85\linewidth]{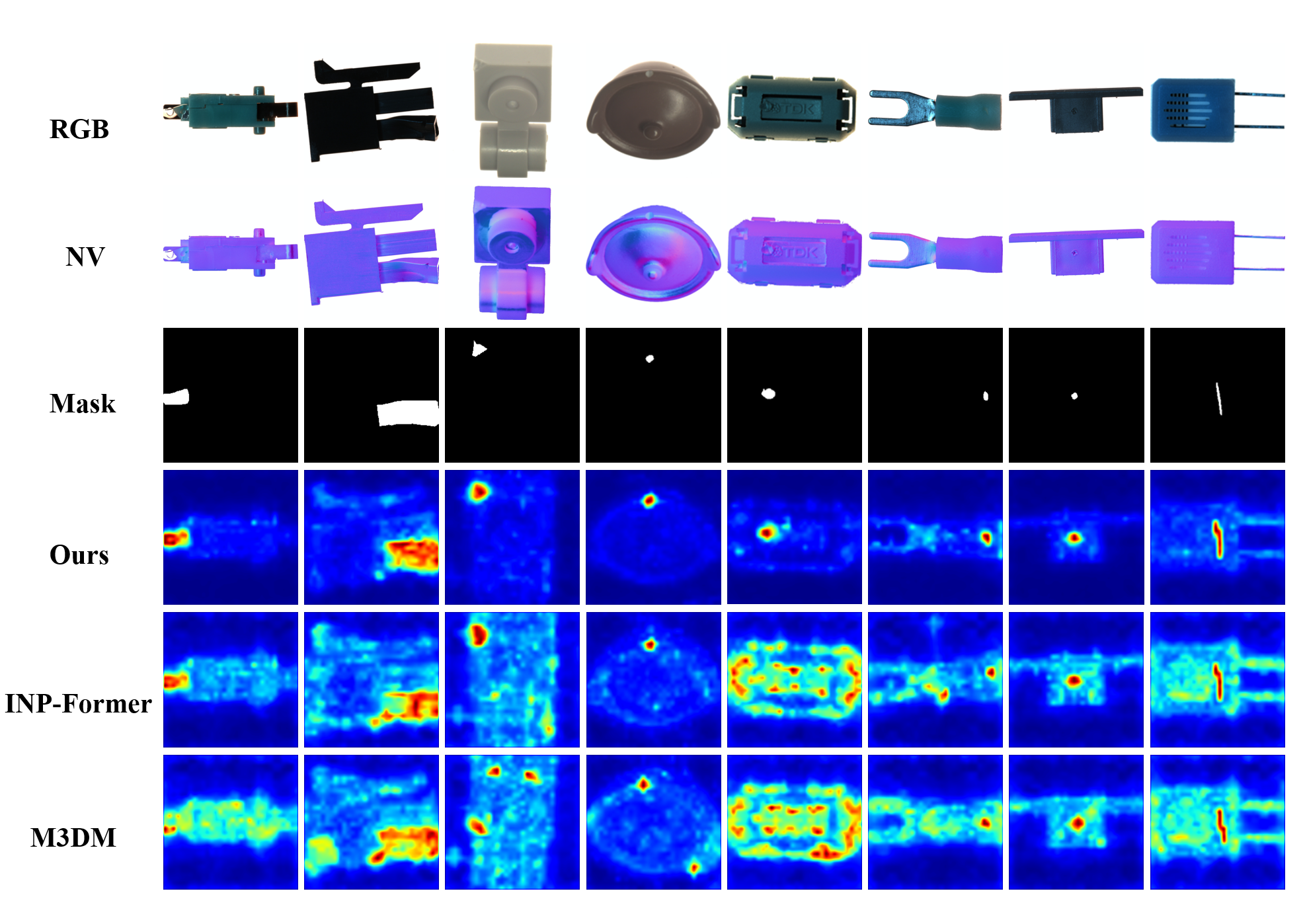} 
\end{center}
   \caption{Qualitative comparison with state-of-the-art methods on various categories from the Real-IAD-MVN dataset (top-down view). From top to bottom: Input RGB, input normal map, ground truth mask, our CPRN (Ours) heatmap, INP-Former~\cite{INP-Former} heatmap, and M3DM~\cite{wang2024m3dm} heatmap. Our method consistently produces more accurate and tightly localized anomaly maps with fewer false positives.}
\label{fig:cmp}
\end{figure*}

\subsection{Modality Comparison: RGB-NV vs. RGB-3D Point Cloud}

Our first experiment (Table \ref{tab:main_results}) investigates the most effective geometric modality in a single-view (top-down) setting for fair alignment. The results are conclusive: 2D+3D methods (e.g., M3DM at 0.841 I-Auroc, 0.830 P-AUPRO) improve over RGB-only (SimpleNet at 0.659 I-Auroc, 0.629 P-AUPRO). However, our CPRN baseline using RGB+NV data achieves a massive leap in performance (0.973 I-Auroc, 0.841 P-AUPRO). This demonstrates that for industrial inspection, the dense, high-fidelity geometric detail from photometric stereo (PS) normals is a far superior data source than sparse 3D point clouds, especially for localizing fine-grained defects (P-AUPRO).

\subsection{Full Multi-View Dataset Evaluation}

Our second experiment (Table 3) evaluates the performance of different single-view methods when applied across the full 5-view dataset. All metrics are computed by treating each view from all samples as an independent image. Here, our CPRN baseline again sets a new state-of-the-art (0.9759 I-Auroc, 0.8558 P-AUPRO), significantly outperforming strong recent competitors like INP-Former~\cite{INP-Former} (0.916 I-Auroc) and Dinomaly~\cite{dinomaly} (0.913 I-Auroc). While the dataset remains challenging, with other methods showing competitive performance on specific categories (e.g., INP-Former~\cite{INP-Former} on 'common\_mode\_filter' and D3M~\cite{zhu2025real} on 'lego\_propeller'), our strong average performance validates the robustness of our unified prototype fusion mechanism for this new, challenging modality.

\begin{table*}[h]
  \centering
  \caption{Ablation study of our CPRN baseline. Performance is evaluated on the full Real-IAD-MVN dataset, with our single-view model applied to all 5 views and results aggregated.}
  \resizebox{0.9\textwidth}{!}{%
    \begin{tabular}{l|ccc|cccc}
        \toprule
    \textbf{Configuration} & \textbf{RGB Stream} & \textbf{NV Stream} & \textbf{Fusion Strategy} & \textbf{I-Auroc} & \textbf{I-F1} & \textbf{P-AUROC} & \textbf{P-AUPRO} \\
    \midrule
    Baseline (RGB only) & \checkmark & & None (Reconstruction) & 0.9152 & 0.9413 & 0.9421 & 0.6410 \\
    Baseline (NV only) & & \checkmark & None (Reconstruction) & 0.9215 & 0.9458 & 0.9503 & 0.6952 \\
    Naive Fusion (Concat) & \checkmark & \checkmark & Concatenation & 0.9521 & 0.9614 & 0.9550 & 0.8104 \\
    Ours (w/o $L_p$) & \checkmark & \checkmark & UCP (Ours) & 0.9712 & 0.9805 & 0.9681 & 0.8420 \\
    \textbf{CPRN (Full Model)} & \checkmark & \checkmark & \textbf{UCP + $L_p$ (Ours)} & \textbf{0.9759} & \textbf{0.9833} & \textbf{0.9707} & \textbf{0.8558} \\
    \bottomrule
    \end{tabular}%
    }
  \label{tab:ablation_study}%
\end{table*}

\subsection{Qualitative and Visual Analysis}
Figure \ref{fig:multiview} illustrates the core advantage of our Real-IAD-MVN dataset: defects on object sides (e.g., 'lego\_pin\_connector\_plate' and 'limit\_switch') are invisible in the top-down view but are clearly revealed in the side-view normal maps, eliminating critical blind spots. Furthermore, Figure \ref{fig:cmp} provides a qualitative comparison on the top-down view, showing our CPRN baseline produces cleaner, more precisely localized heatmaps than SOTA methods like INP-Former~\cite{INP-Former} and M3DM~\cite{wang2024m3dm}, which exhibit more diffuse, false-positive activations. This suggests our unified prototype approach learns a more constrained and accurate representation of normality.

\subsection{Ablation Study}
We validate our CPRN baseline's design on the full 5-view dataset in Table \ref{tab:ablation_study}. The study shows a clear progression: the 'NV only' (multi-view normal) reconstruction (0.7952 P-AUPRO) performs notably better than 'RGB only' (0.641 P-AUPRO), confirming the geometric stream's power. A 'Naive Fusion (Concat)' baseline (0.8104 P-AUPRO) improves upon single modalities, proving fusion is beneficial. The major performance leap comes from our 'UCP (Ours)' module (0.8420 P-AUPRO), which replaces naive concatenation, demonstrating the superiority of learning unified prototypes. Finally, adding the Prototype Distribution Entropy Regularization ($L_p$) provides the last significant boost to our final performance (0.8558 P-AUPRO), confirming its importance in preventing prototype collapse and ensuring a diverse set of normal concepts.

\section{Conclusion}
\label{sec:conclusion}

In this paper, we introduced Real-IAD-MVN, a new large-scale industrial anomaly detection dataset. Its primary contribution is the introduction of high-fidelity, dense surface normal maps captured from five distinct viewpoints, replacing traditional sparse 3D point cloud data. We demonstrated that this new multi-view normal (PS-3D) modality is a far superior data source for detecting subtle, geometric defects compared to both RGB-only and RGB + 3D point cloud approaches.
To validate this new dataset, we proposed a strong baseline, the Cross-modal Prototype Reconstruction Network (CPRN). This model leverages the concept of unified prototypes to effectively leverages the joint distribution of normality.

\section*{Acknowledgements}  
 This research was supported by the Shanghai Agricultural Science and Technology Project (grant number T20252016), and the Shanghai Science and Technology Project (grant number 24YF2716900) and Suzhou Major Project (“Jiebang Guashuai”) for Transformation of Scientific and Technological Achievements (Grant No. SZC2024020).

{
    \small
    \bibliographystyle{ieeenat_fullname}
    \bibliography{main}
}
\clearpage
\appendix
\renewcommand\thefigure{A\arabic{figure}}
\renewcommand\thetable{A\arabic{table}}  
\renewcommand\theequation{A\arabic{equation}}
\setcounter{equation}{0}
\setcounter{table}{0}
\setcounter{figure}{0}

\section{Generative Potential of Real-IAD-MVN}
\label{sec:appendix_generative}

To showcase the richness of the learned representation in our new dataset and to explore future research directions, we present an \textbf{Awareness-Guided Generative Module}. This module is designed to validate the deep, intrinsic understanding of 3D structure that can be learned from our high-fidelity 2D (RGB) and Pseudo-3D (Normal Vector) data.

Its purpose is to generate a high-quality geometric representation (e.g., a single-channel depth image) using only the RGB and normal vector (NV) inputs provided by our Real-IAD-MVN dataset. The success of this complex cross-modal generation task serves as compelling evidence that our new data modality (MV-RGB + MV-NV) captures a profound understanding of object structure, potentially enabling hardware-agnostic applications where expensive 3D scanners can be replaced.

Concretely, the module follows a two-stage pipeline conditioned by a hierarchical fusion mechanism. First, RGB and NV features are encoded and fused by a \textbf{Hierarchical Dual Conditioner (HDC)}. The HDC performs multi-level cross-modal interaction and injects the resulting conditional features into a \textbf{Denoising Unet}, which predicts a latent geometric representation through a conditional diffusion process. Second, the denoised latent code is forwarded to an \textbf{Adversarial Dense Generation} decoder that reconstructs a high-fidelity depth image. The overall design therefore combines cross-modal prototype-aware conditioning, diffusion-based latent generation, and adversarial dense decoding in a unified generation framework.
\subsection{Two-Stage Generation for High-Fidelity Depth}
We first generate a latent geometric representation using diffusion, then decode it into a dense depth image using an adversarial network.

\textbf{1. Diffusion-based Latent Generation.} We adopt a conditional denoising diffusion probabilistic model (DDPM) to generate a latent representation of the geometry. A forward process gradually adds Gaussian noise to a ground truth latent vector, and a \textbf{Denoising Unet} is trained to reverse this process. The reverse process is conditioned on the fused features from the RGB image ($I$) and the pseudo-3D normal vectors ($P$):
\begin{equation}
    p_\theta(x_{t-1}|x_t, I, P) = \mathcal{N}(x_{t-1}; \mu_\theta(x_t, t, I, P), \sigma_t^2\mathbf{I})
\end{equation}
The Unet, $\epsilon_\theta$, is trained to predict the noise added at each timestep $t$:
\begin{equation}
    \mathcal{L}_\text{diffusion} = \mathbb{E}_{t,x_0,\epsilon}\left[\|\epsilon - \epsilon_\theta(x_t, t, I, P)\|_2^2\right]
\end{equation}

\textbf{2. Adversarial Dense Generation.} The denoised latent representation from the Unet is passed to a final \textbf{Decoder} network, which generates the high-resolution, single-channel depth image $\hat{I}_{\text{depth}}$. This stage is trained with a combination of a reconstruction loss and an adversarial loss to ensure high fidelity and realism:
\begin{align}
    \mathcal{L}_\text{reconstruct} &= \mathbb{E}\left[\|\hat{I}_{\text{depth}} - I_{\text{depth\_GT}}\|_2^2\right] \\
    \mathcal{L}_\text{adv} &= \mathbb{E}\left[\log(1-D(\hat{I}_{\text{depth}}))\right] + \mathbb{E}\left[\log D(I_{\text{depth\_GT}})\right]
\end{align}
where $D$ is a discriminator network distinguishing generated images from ground truth (GT) depth images.

\subsection{Hierarchical Conditioning for Input Fusion}
The conditioning of the Denoising Unet is orchestrated by a \textbf{Hierarchical Dual Conditioner (HDC)}, which effectively fuses the image and normal vector modalities. It uses resolution-aware information routing, extracting features $F_{\mathcal{M}}^l$ at $L$ hierarchical levels. The core is a Bi-Modal Cross-Attention (BCA) module that uses features from the Unified Cross-Modal Prototype Extraction (UCP) module (detailed in our main paper's baseline) as a rich feature source.
\begin{equation}
    \text{CA}_{\mathcal{M}}(Q, \tilde{F}_{\mathcal{M}}^l) = \text{Softmax}\left(\frac{QW_Q^{\mathcal{M}} (W_K^{\mathcal{M}}\tilde{F}_{\mathcal{M}}^l)^T}{\sqrt{d_k}}\right)W_V^{\mathcal{M}}\tilde{F}_{\mathcal{M}}^l
\end{equation}
To dynamically balance the modalities, a weighting coefficient $\alpha_l$ is computed for each level $l$, based on the estimated mutual information $\mathcal{I}$ that each modality's features provide with respect to the Unet's query features $Q$:
\begin{equation}
    \alpha_l = \sigma\left(\beta_l \cdot \frac{\mathcal{I}(Q; \tilde{F}_P^l)}{\mathcal{I}(Q; \tilde{F}_P^l) + \mathcal{I}(Q; \tilde{F}_I^l)}\right)
\end{equation}
The final conditional features injected into the Unet are a weighted sum of the attention outputs, ensuring that the most relevant information from each modality (RGB and NV) is used at the most appropriate stage of generation.

\begin{figure}[t]
  \centering
  \includegraphics[width=0.5\textwidth]{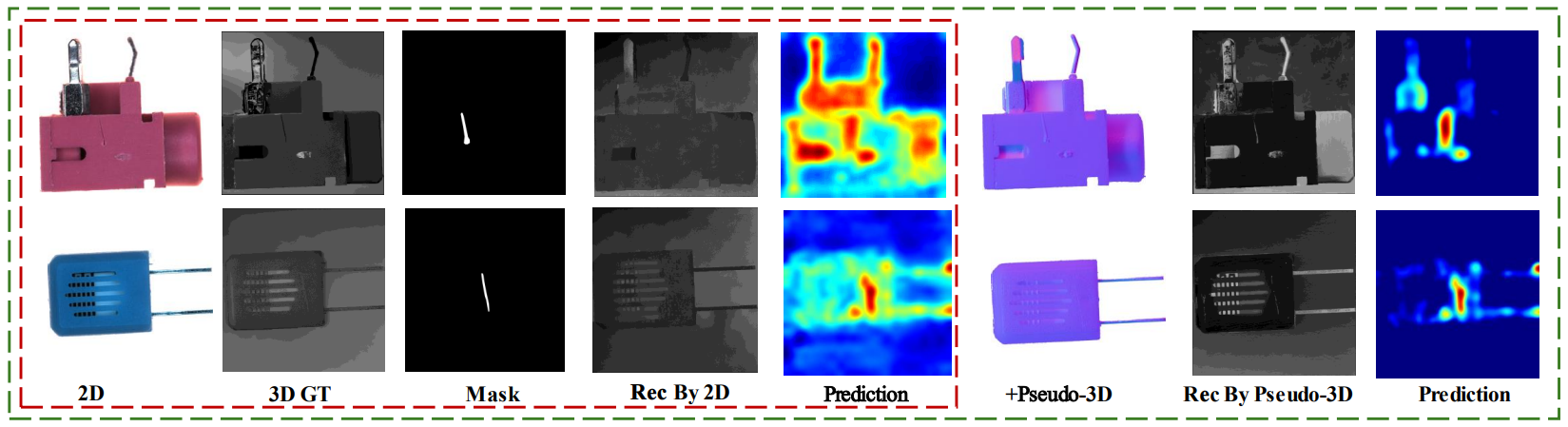}
  \caption{Qualitative comparison of anomaly segmentation from the generative module's output on the Real-IAD D³ dataset. The right panel (+Pseudo-3D) shows that generating geometry from normal vectors (Rec By Pseudo-3D) leads to a sharper, more accurate anomaly prediction compared to the 2D-only baseline (left panel).}
\label{fig:gen_compare}
\end{figure}

\begin{figure}[t]
  \centering
  \includegraphics[width=0.5\textwidth]{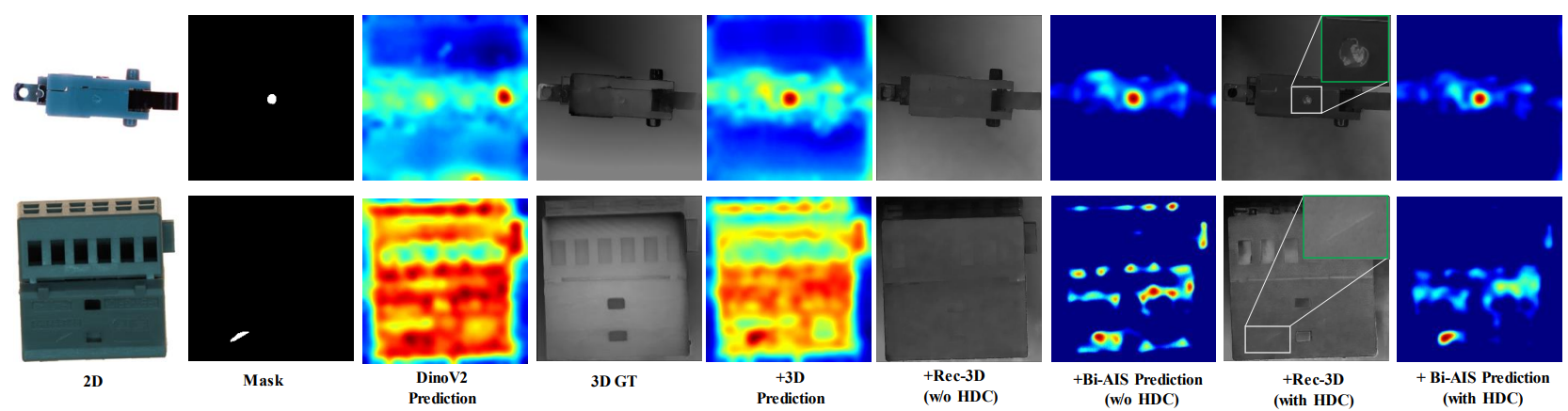}
\caption{Qualitative results of the generative module's ablation study on Real-IAD D³. The baseline 2D prediction (DinoV2) is diffuse. Adding 3D reconstruction (+Rec-3D) improves focus. The full model (+Bi-AIS Prediction (with HDC)) demonstrates the sharpest and most accurate localization by effectively fusing the RGB and normal vector data.}
\label{fig:gen_ablation}
\end{figure}
\subsection{Qualitative Analysis of Generative Module}
While the quantitative results of this generative model are reserved for future work, we provide qualitative results here to demonstrate its effectiveness, which further validates the quality of our Real-IAD-MVN data.

The qualitative results in Figure \ref{fig:gen_compare} (conducted on the similar Real-IAD D³ dataset) visually demonstrate the principle of defect enhancement. The left panel shows the prediction from a 2D-only baseline, which produces a diffuse and inaccurate anomaly map. In contrast, the right panel shows that after generating a geometric representation guided by pseudo-3D information (normal vectors), the model's reconstruction is sharper, and the final prediction is precisely focused on the true anomalous regions (e.g., the bent pins), providing a much clearer signal for detection.

The progressive improvement in localization accuracy is also visualized in the ablation study in Figure \ref{fig:gen_ablation}. The baseline model's (DinoV2) heatmap is noisy and unfocused. Adding the reconstruction module with pseudo-3D data (+Rec-3D) begins to consolidate the activation. Finally, the full model with the Hierarchical Dual Conditioner (+Bi-AIS Prediction (with HDC)) produces a remarkably clean and accurate anomaly map, demonstrating the critical role of sophisticated fusion of the RGB and normal vector data.

\section{Real-IAD-MVN Dataset Visualizations}
\label{sec:appendix_vis}

\begin{figure*}[t]
\begin{center}
\includegraphics[width=0.8\linewidth]{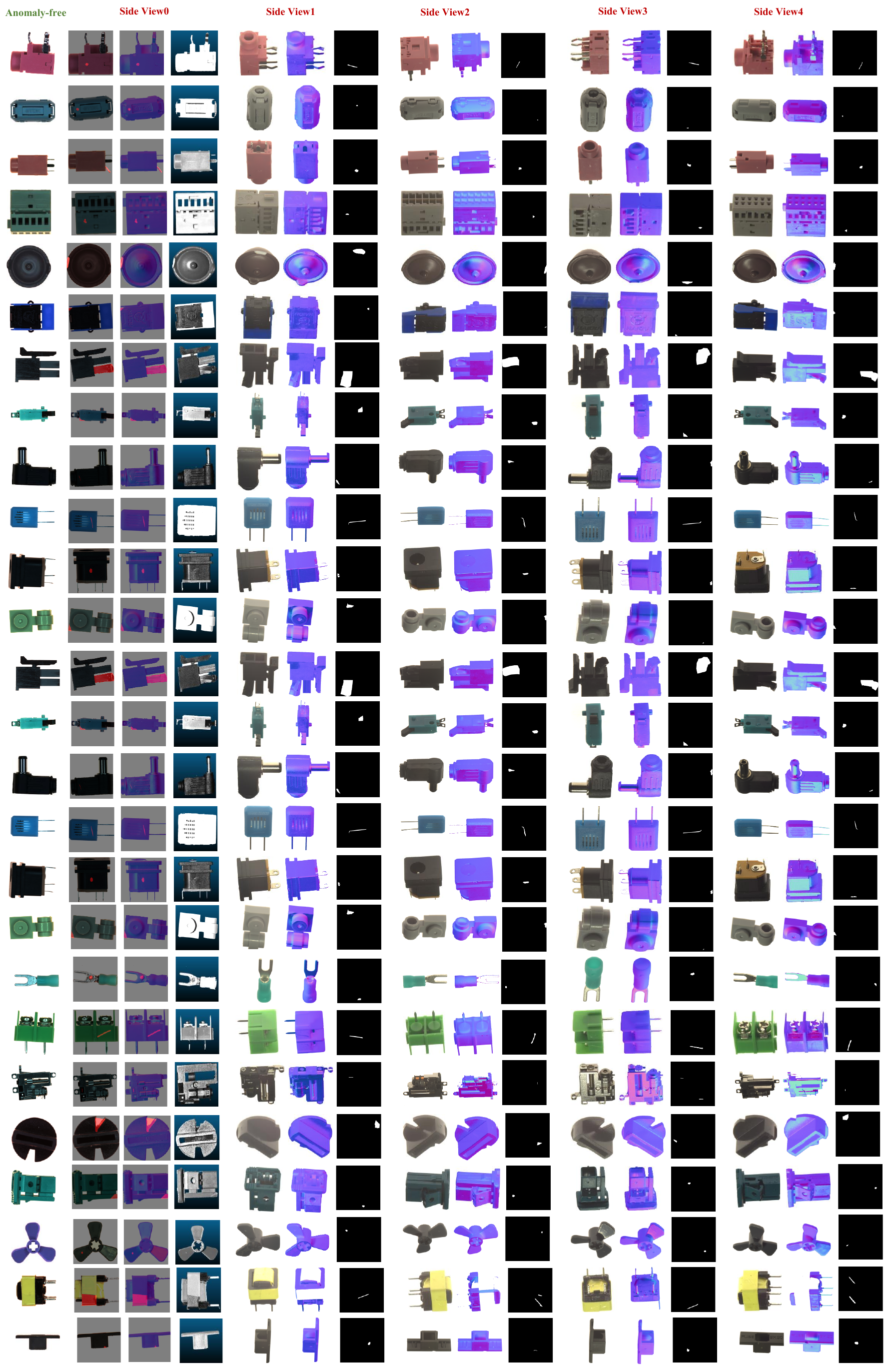}
\end{center}
   \caption{Comprehensive visualization of all 20 categories in the Real-IAD-MVN dataset. Each row corresponds to one object category. The first column shows the anomaly-free top-down RGB image. The subsequent columns (Side View 0-4) display the five captured viewpoints. For each viewpoint, we show the RGB image, its corresponding high-fidelity PS Normal Map (NV), and the pixel-level ground truth anomaly mask.}
\label{fig:multiview_all}
\end{figure*}

To provide a comprehensive overview of the data included in Real-IAD-MVN, Figure \ref{fig:multiview_all} visualizes all 20 object categories. For each category, we display the anomaly-free top-down view, followed by all five distinct, calibrated viewpoints (Side View 0 through Side View 4). Each viewpoint is represented by its (1) RGB image, (2) high-fidelity PS Normal Vector (NV) map, and (3) the corresponding ground truth anomaly mask (shown in white, if an anomaly is present in that view).

\clearpage
\section{Camera Calibration Parameters}
\label{sec:appendix_calibration}

In Section 3.1 of our main paper, we describe our
integrated multi-view photometric stereo (MVPS) gantry,
which operates with five pre-calibrated viewpoints. To en-
able and encourage future research in multi-view 3D recon-
struction, geometric registration, and novel view synthesis
(e.g., NeRF-based tasks), we provide the full calibration pa-
rameters for four of our primary acquisition cameras below.
The parameters are provided in the `.yml` format, compati-
ble with standard calibration tools.

\subsection{Camera 1 Parameters}
\begin{verbatim}
calibration_file_ver: 3.1
calibration_ver: 3.1.359264
calibration_time: 2025-1-24 15:6:57
calibration_type: MVP_CALIB_TYPE_NORMAL
frame_num: 1
image_width: 4096
image_height: 3000
board_width: 0
board_height: 0
board_dx: 2.0000000000000000e+000
board_dy: 2.0000000000000000e+000
points_num: 0
H_matrix: 
  rows: 3
  cols: 3
  data: 
    - 7.5142908289968972e+001
    - 4.9162084131020807e+000
    - 8.2109891582446537e+001
    - -1.7980084333606616e+000
    - 5.7724585197937522e+001
    - 1.0859333933073324e+003
    - 8.2071064611873192e-005
    - 2.3720123199760068e-003
    - 1.0000000000000000e+000
distortion_model: 3
distortion_level: 2
distortion_matrix: 
  rows : 12
  cols : 1
  data: 
    - -7.0994615634284253e-001
    - 1.5565761184853105e+001
    - -6.7707702880165407e-002
    - -1.2152307380140845e-002
    - -1.1598420108419958e+002
    - 0.0000000000000000e+000
    - 0.0000000000000000e+000
    - 0.0000000000000000e+000
    - 0.0000000000000000e+000
    - 0.0000000000000000e+000
    - 0.0000000000000000e+000
    - 0.0000000000000000e+000
intrinsic_matrix: 
  rows : 3
  cols : 3
  data: 
    - 1.9426147574516781e+004
    - 0.0000000000000000e+000
    - 1.9428998682433419e+003
    - 0.0000000000000000e+000
    - 2.0040639975377839e+004
    - -1.4103926763176123e+003
    - 0.0000000000000000e+000
    - 0.0000000000000000e+000
    - 1.0000000000000000e+000
extrinsic_matrix: 
  rows : 3
  cols : 4
  data: 
    - 9.9953780819267646e-001
    - 4.1006888450079194e-003
    - -3.0122323024239783e-002
    - -2.4804538528430975e+001
    - -2.1737068720812217e-002
    - 7.8910936074326898e-001
    - -6.1386799609588383e-001
    - 3.2255977192006476e+001
    - 2.1252525421862352e-002
    - 6.1423904234291860e-001
    - 7.8883377781687780e-001
    - 2.5895272008921592e+002
\end{verbatim}

\subsection{Camera 2 Parameters}
\begin{verbatim}
calibration_file_ver: 3.1
calibration_ver: 3.1.359264
calibration_time: 2025-1-24 15:4:51
calibration_type: MVP_CALIB_TYPE_NORMAL
frame_num: 1
image_width: 4096
image_height: 3000
board_width: 0
board_height: 0
board_dx: 2.0000000000000000e+000
board_dy: 2.0000000000000000e+000
points_num: 0
H_matrix: 
  rows: 3
  cols: 3
  data: 
    - 7.4755050461534353e+001
    - 6.2458574231518309e+000
    - 1.0981382636031776e+003
    - -1.6285738267351551e+000
    - 5.7629797788179566e+001
    - 1.7663434952193629e+002
    - -4.8201648078641062e-004
    - 2.8343758580036344e-003
    - 1.0000000000000000e+000
distortion_model: 3
distortion_level: 2
distortion_matrix: 
  rows : 12
  cols : 1
  data: 
    - -2.4410037940517855e+000
    - 4.3015853395902596e+001
    - 2.1299813348685793e-002
    - 8.8716663242255092e-002
    - -3.5257553640216827e+002
    - 0.0000000000000000e+000
    - 0.0000000000000000e+000
    - 0.0000000000000000e+000
    - 0.0000000000000000e+000
    - 0.0000000000000000e+000
    - 0.0000000000000000e+000
    - 0.0000000000000000e+000
intrinsic_matrix: 
  rows : 3
  cols : 3
  data: 
    - 1.9798511319027260e+004
    - 0.0000000000000000e+000
    - -4.3049246222583417e+002
    - 0.0000000000000000e+000
    - 2.0271741030735127e+004
    - 2.4744169291864846e+003
    - 0.0000000000000000e+000
    - 0.0000000000000000e+000
    - 1.0000000000000000e+000
extrinsic_matrix: 
  rows : 3
  cols : 4
  data: 
    - 9.9188951598251840e-001
    - 9.9339040733826808e-002
    - 7.9290245743398075e-002
    - 2.0339137531916062e+001
    - -5.6639905656809125e-003
    - 6.5775238666472491e-001
    - -7.5321292942150830e-001
    - -2.9859351241362621e+001
    - -1.2697679825400085e-001
    - 7.4665490879183360e-001
    - 6.5298035183474057e-001
    - 2.6342833350187118e+002
\end{verbatim}

\subsection{Camera 3 Parameters}
\begin{verbatim}
calibration_file_ver: 3.1
calibration_ver: 3.1.359264
calibration_time: 2025-1-24 15:54:51
calibration_type: MVP_CALIB_TYPE_NORMAL
frame_num: 1
image_width: 4096
image_height: 3000
board_width: 0
board_height: 0
board_dx: 2.0000000000000000e+000
board_dy: 2.0000000000000000e+000
points_num: 0
H_matrix: 
  rows: 3
  cols: 3
  data: 
    - 7.5368561341388229e+001
    - 6.6572215304853710e+000
    - 2.2010105906328986e+003
    - 1.2956815846651235e+000
    - 5.6687607049282498e+001
    - 4.9215373641409366e+002
    - 1.3453776666924063e-004
    - 3.0541386844316651e-003
    - 1.0000000000000000e+000
distortion_model: 3
distortion_level: 2
distortion_matrix: 
  rows : 12
  cols : 1
  data: 
    - 1.8004248377403349e-002
    - -3.0044854730835597e+000
    - -2.5340313073236019e-003
    - 1.5663430597834375e-002
    - 3.9016750900379250e+001
    - 0.0000000000000000e+000
    - 0.0000000000000000e+000
    - 0.0000000000000000e+000
    - 0.0000000000000000e+000
    - 0.0000000000000000e+000
    - 0.0000000000000000e+000
    - 0.0000000000000000e+000
intrinsic_matrix: 
  rows : 3
  cols : 3
  data: 
    - 1.7313015180332859e+004
    - 0.0000000000000000e+000
    - 2.9813759579949819e+003
    - 0.0000000000000000e+000
    - 1.7258484049521408e+004
    - 1.2130817833681663e+003
    - 0.0000000000000000e+000
    - 0.0000000000000000e+000
    - 1.0000000000000000e+000
extrinsic_matrix: 
  rows : 3
  cols : 4
  data: 
    - 9.9940303465816926e-001
    - -3.2638864426226488e-002
    - -1.1326025119533373e-002
    - -1.0403174684703622e+001
    - 1.5144923326588894e-002
    - 7.0855235884226586e-001
    - -7.0549570237988890e-001
    - -9.6411733094746737e+000
    - 3.1051660398014748e-002
    - 7.0490301411472722e-001
    - 7.0862369074036691e-001
    - 2.3080255579353312e+002
\end{verbatim}

\subsection{Camera 4 Parameters}
\begin{verbatim}
calibration_file_ver: 3.1
calibration_ver: 3.1.359264
calibration_time: 2025-1-24 15:56:5
calibration_type: MVP_CALIB_TYPE_NORMAL
frame_num: 1
image_width: 4096
image_height: 3000
board_width: 0
board_height: 0
board_dx: 2.0000000000000000e+000
board_dy: 2.0000000000000000e+000
points_num: 0
H_matrix: 
  rows: 3
  cols: 3
  data: 
    - 7.2093441984252976e+001
    - 6.3939899920335943e+000
    - 1.2086169768470568e+003
    - 9.8103425121876583e-003
    - 5.3897464619363276e+001
    - 1.2205660967156343e+003
    - 3.1503871509957305e-004
    - 2.7843508343726778e-003
    - 1.0000000000000000e+000
distortion_model: 3
distortion_level: 2
distortion_matrix: 
  rows : 12
  cols : 1
  data: 
    - 1.8131330524292905e-001
    - -1.2469765989270881e+001
    - 6.3753197148693971e-003
    - 9.0223021157360868e-003
    - 1.5616130476247724e+002
    - 0.0000000000000000e+000
    - 0.0000000000000000e+000
    - 0.0000000000000000e+000
    - 0.0000000000000000e+000
    - 0.0000000000000000e+000
    - 0.0000000000000000e+000
    - 0.0000000000000000e+000
intrinsic_matrix: 
  rows : 3
  cols : 3
  data: 
    - 1.7832381835705353e+004
    - 0.0000000000000000e+000
    - 3.6234241894269994e+003
    - 0.0000000000000000e+000
    - 1.7843555944438864e+004
    - 1.0788558390330131e+003
    - 0.0000000000000000e+000
    - 0.0000000000000000e+000
    - 1.0000000000000000e+000
extrinsic_matrix: 
  rows : 3
  cols : 4
  data: 
    - 9.9686931222851527e-001
    - -5.1912991967135176e-002
    - -5.9637367497796648e-002
    - -3.3927864116395583e+001
    - -4.6345694251529318e-003
    - 7.1460287931792632e-001
    - -6.9951500744213679e-001
    - 1.9897718916501284e+000
    - 7.8930951491100951e-002
    - 6.9760143788236939e-001
    - 7.1212508645683870e-001
    - 2.5054365609050163e+002
\end{verbatim}


\end{document}